\pdfoutput=1
\documentclass[11pt, dvipsnames, table]{article}

\usepackage{acl}

\usepackage{times}
\usepackage{latexsym}
\usepackage[T1]{fontenc}
\usepackage[utf8]{inputenc}
\usepackage{microtype}
\usepackage{inconsolata}
\usepackage{amsmath}
\usepackage{amssymb}
\usepackage{amsthm}
\usepackage{graphicx}
\usepackage{mathtools}
\usepackage{listings}
\usepackage{tikz}
\usepackage{siunitx}
\usepackage{tipa}
\usepackage{float}
\usepackage{bbm}
\usepackage{breqn}
\usepackage{utils}
\usepackage{tikz-dependency}
\usepackage{url}
\usepackage{caption}
\usepackage{subcaption}
\usepackage{expex}

\usepackage{todonotes}

\makeatletter
\newcommand*\iftodonotes{\if@todonotes@disabled\expandafter\@secondoftwo\else\expandafter\@firstoftwo\fi}  % defines \iftodonotes{<true>}{<false>}, thanks to https://tex.stackexchange.com/questions/126559/conditional-based-on-packageoption
\makeatother

\definecolor{blu}{HTML}{3468C0}
\definecolor{yello}{HTML}{e6ab02}
\definecolor{piink}{HTML}{e7298a}
\definecolor{greeen}{HTML}{66a61e}
\definecolor{reed}{HTML}{b2182b}
\definecolor{firebrick}{HTML}{B22222}
\definecolor{steelblue}{HTML}{4682B4}
\definecolor{puuurpl}{HTML}{B5739D}
\definecolor{gren}{HTML}{1b9e77}
\definecolor{prpl}{HTML}{7570b3}
\definecolor{orang}{HTML}{d95f02}
\definecolor{vlightgreen}{HTML}{82B366}
\definecolor{vviolet}{HTML}{56517E}

\definecolor{regexv1}{HTML}{B46504}
\definecolor{regexv2}{HTML}{56517E}
\definecolor{regexv3}{HTML}{0E8088}

\definecolor{naanpollution}{HTML}{488795}
\definecolor{ananpollution}{HTML}{ae4d9f}

\usepackage{booktabs, multirow}

\Crefname{figure}{{Fig.}}{{Figs.}}
\crefname{section}{§}{§§}
\Crefname{section}{§}{§§}
\Crefname{appendix}{{App.}}{{Apps.}}

\title{\texttt{semantic-features}: A User-Friendly Tool for Studying Contextual Word Embeddings in Interpretable Semantic Spaces}

\author{Jwalanthi Ranganathan$^1$\quad Rohan Jha$^1$ \quad Kanishka Misra$^{1,2,\bigstar}$ \quad Kyle Mahowald$^1$\\
$^1$The University of Texas at Austin $^2$Toyota Technological Institute at Chicago\\
\texttt{\{jwalanthi,rjha,kyle\}@utexas.edu \quad \{kanishka\}@ttic.edu}}

\begin{document}

\maketitle

{\let\thefootnote\relax\footnotetext{\hspace{-0.2cm}$^{\bigstar}$Work partly done at UT-Austin before joining TTIC.}}

\section{Introduction}

The advent of distributional semantic embeddings has enabled major progress in the computational understanding of word meaning by enabling precise statistical explorations of semantic spaces
\citep{erk-2009-representing,mikolov-etal-2013-efficient, pennington2014glove}.
More recently, the rise of LMs have made it possible to study embeddings of words in \textit{context}.
\citet{chronis-etal-2023-method} developed a method for projecting \textit{contextual word embeddings (CWEs)} into a interpretable semantic feature space defined by one of three different semantic norms \citep
{binder_toward_2016, buchanan_2019, mcrae2005semantic}. 
This is achieved by training feed-forward models which map from CWEs from BERT to a vector whose values correspond to feature norms.

Our goal in this paper is twofold: first, we
introduce \texttt{semantic-features}\footnote{ \texttt{semantic-features} is available at \url{https://github.com/jwalanthi/semantic-features}} as
% give 
an extensible, easy-to-use library based on \citet{chronis-etal-2023-method} for studying word embeddings from any LM in context.
Second, we show its ease of use through an online application which researchers can use without additional programming. We demonstrate these tools with a linguistic experiment that uses this method to measure the contextual effect of the choice of dative construction (prepositional or double object) on the semantic interpretation of utterances.

The dative construction has been of particular interests to theoretical \citep{goldberg1995constructions, hovav2008english, beavers2011aspectual} and computational linguists \citep{bresnan2007predicting, hawkins-etal-2020-investigating, liu-wulff-2023-development, jumelet-etal-2024-language, misra2024generating, yao2025both}. This is primarily due to its several interesting properties such as its participation in alternation behavior \citep{levin1993english}, flexible interpretation of the event it describes---caused motion vs. caused possession \citep{goldberg1995constructions, hovav2008english, beavers2011aspectual}, and interesting feature specific preferences that humans demonstrate while choosing between two dative constructions during production \citep{bresnan2007predicting}.

\begin{figure}[!t]
    \centering
    \includegraphics[width=\linewidth]{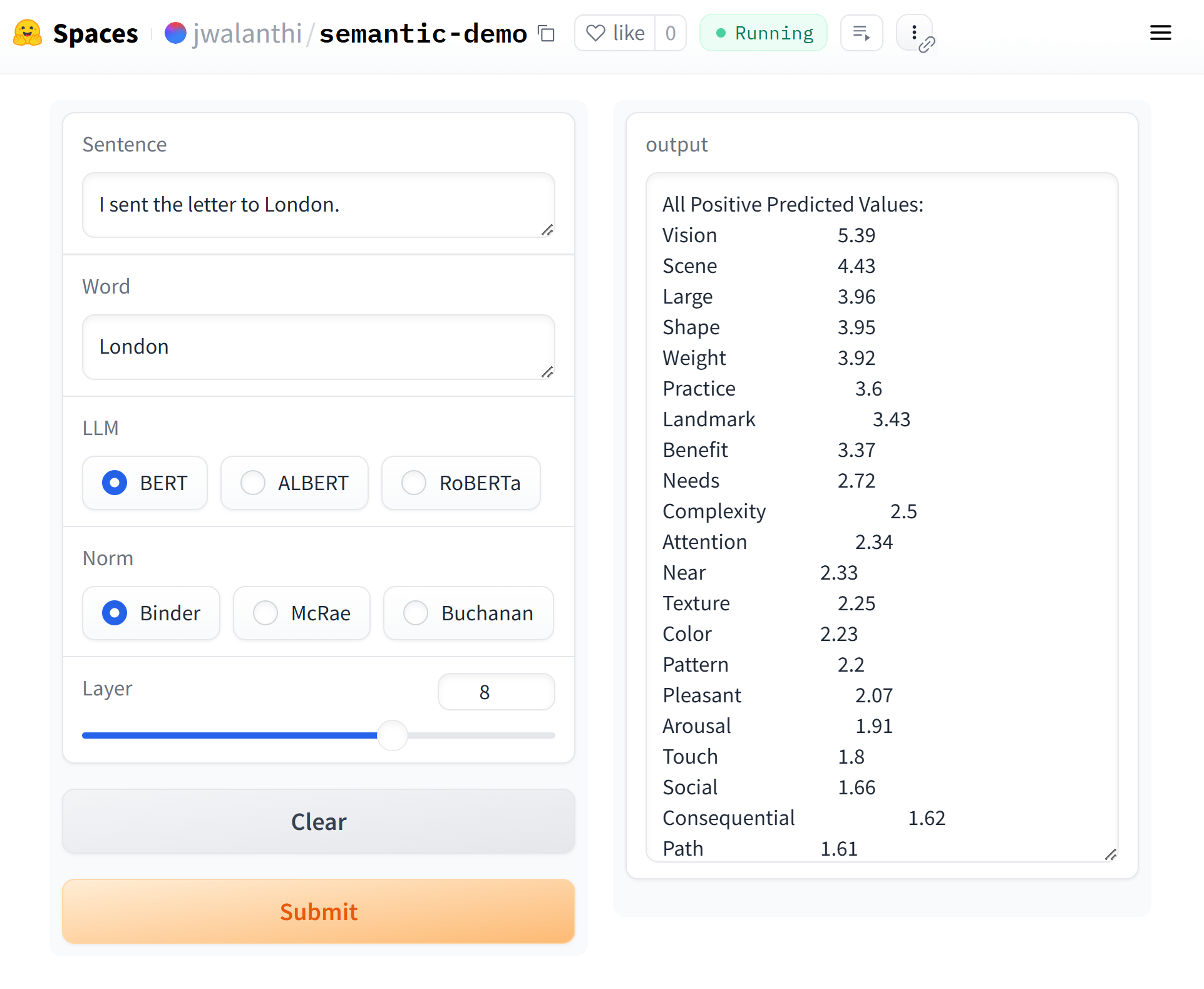}
    \caption{Interactive Demo in Use}
    \label{fig:demo}
    \vspace{-1em}
\end{figure}
% Second, to provide an online demo version which can be used without additional programming.
% We give a proof-of-concept demonstration of how this method can be used for a linguistic experiment measuring the contextual effect of the choice of dative construction (prepositional or double object) on the semantic interpretation of utterances \citep{bresnan2007predicting}.

Our case study focuses on the semantics of the arguments of the dative construction---in particular its \textit{recipient} argument \citep{beavers2011aspectual, petty2022language}.
Specifically, we hypothesize that ``London'' in ``I sent \textbf{London} the letter.'' (\textit{double object}; DO) should be more likely to be interpreted as an animate referent (e.g., as the name of a person) than in ``I sent the letter to \textbf{London}.'' (\textit{prepositional object}; PO) This is because the DO dative is more canonically associated with possession transfer events, which constrains the recipient to be animate \citep{beavers2011aspectual}. The PO dative, on the other hand, is associated with both possession transfer and `caused-motion' \citep{goldberg1995constructions} and allows for inanimate recipients.
We test whether LMs learn this distinction by projecting the embedded representation from the token ``London'' into a more interpretable semantic space and analyze it for animate vs. inanimate features.
We include a full demonstration of how to easily obtain such measures from models that have already been trained, in addition to describing our full system for training projections from scratch.

\section{\texttt{semantic-features}}
Our extensible system for training models and analyzing embeddings performs three main tasks: embedding extraction, model training, and hyperparameter tuning. 
% We also release a user-friendly tool on HuggingFace that can be directly used, without any programming experience required, to obtain contextual word embeddings.
Below, we summarize our methodology; more details can be found in the README.

\paragraph{Embedding Extraction} The first step is preparing the CWEs which serve as the `source' for the training data. Given a (user-provided) corpus and an LM whose weights/embeddings are accessible, \texttt{semantic-features} extracts an embedding for each word in the corpus using \texttt{minicons} \citep{misra2022minicons}.
We average the embeddings across all contexts to obtain one vector per word, as in \citet{chronis-etal-2023-method}.
While any LM can theoretically serve as the source for word embeddings, autoregressive LMs like GPT-2 are not well-suited for this application because their embeddings only capture left-context for a given word.\footnote{As a note, while \texttt{semantic-demo} provides the ability to train these MLPs, it does not provide the raw training data itself. Users must obtain corpus data and feature norm data from their respective sources.}

\paragraph{Model training} All models use a multi-layer perceptron (MLP) to perform feature prediction. All hyperparameters can be user-specified except for the MLP architecture. While \citet{chronis-etal-2023-method} experimented with other architectures, we choose MLPs to maintain a fully neural system end-to-end. Models are trained with a 80-20 train-validation split, and loss is calculated as mean-squared error between the predicted vector and the ground-truth feature-norm vector. 

\paragraph{Hyperparameter tuning} 
Our system allows for hyperparameter tuning by using
\texttt{optuna} \citep{akiba2019optunanextgenerationhyperparameteroptimization}. We specifically use the \texttt{TPESampler} module, which searches for the combination of hyperparameters which minimizes validation loss using a Tree-Structured Parzen Estimator algorithm. \texttt{optuna} searches for the optimal values for hidden size, batch size, and learning rate over a specified set of ranges
in Table \ref{table:optuna}. 
If enabled, the \texttt{MedianPruner} is used to determine which trials to prune. After running 100 trials, the model with the lowest validation loss is saved. 

\paragraph{Interactive Demo}
An interactive demonstration of a selection of models trained using \verb|semantic-features| is available on HuggingFace Spaces as a Gradio app,\footnote{Available at \url{https://huggingface.co/spaces/jwalanthi/semantic-demo}} shown in Figure \ref{fig:demo}. Users can retrieve a model which maps from the CWE of a user-specified word in context from any layer of BERT \citep{devlin-etal-2019-bert}, RoBERTa \citep{liu_roberta_2019}, or ALBERT \citep{lan2019albert} to any of the three semantic feature spaces used by \citet{chronis-etal-2023-method}.\footnote{While these are the models we currently focus on, in principle this can be applied on any masked language model.} The models were trained using the British National Corpus as the source text. For each word that has a pre-defined feature vector, \texttt{semantic-features} extracts the embeddings in each context provided by BNC, averaging the embeddings per word across all contexts. This serves as the source for training, and the feature vector itself serves as the target. 
Further training details, including hyperparameter specification and GPU training hours, are provided in Appendix B. 
The output of the demo is a list of the predicted features sorted greatest to least.

\section{A Small Case Study on Recipient Semantics in Dative Constructions}

Using the tools developed in the previous section, we ask if LMs are sensitive to context-dependent semantics in linguistic constructions. Consider the dative alternation: some ditransitive verbs can take two different argument structures. The first is the %V-NP-NP or 
\textit{double object (DO)} construction and the second is %the V-NP-PP or 
\textit{prepositional object (PO)} construction.
\pex[labeltype=alpha]<london>
\vtop{\labels\halign{\tl #\hfil&& \quad #\hfil\cr
<DO>& I sent \textbf{London} the letter.& \textit{DO}\cr
<PO>& I sent the letter to \textbf{London}.& \textit{PO}\cr
}}
\xe
While both 
are near synonymous, they apply different contextual constraints on their arguments. For instance, in the PO, \textit{London} takes on its ``standard'' definition as an inanimate place/location, but in the DO, it seems that \textit{London} is 
an animate recipient \citep{beavers2011aspectual, hovav2008english}. 
To what extend do LMs learn this distinction? To test this, we project embeddings from LMs to the Binder features \citep{binder_toward_2016} space. We choose the Binder Norms here specifically because each feature has a concrete definition provided by the researchers, which can allow for finer grained person-hood vs. place-hood distinction. We use these definitions (reproduced in Table \ref{tab:binder_feats}) to identify Binder features which capture place-hood (Landmark and Scene) and person-hood (Biomotion, Body, Human, Face, and Speech) to reflect the two possible salient readings. Higher values for a Binder feature from the projected embedding is taken to mean greater activation of the specific feature. 
We choose features which capture person-hood and place-hood distinctively, not those which are applicable for both readings. For example, the Vision feature, which is defined as ``something that you can easily see," can be activated in both contexts, and is therefore not included in either category. We then extract the embeddings for the recipient word each layer of the LM in each context and project them to the Binder space, observing changes in the relevant features. Table \ref{table:london} shows an example set of predictions for (1) using BERT layer 8. We see that, consistent with our predictions, ``London'' is construed as more person-like in the DO and more place-like in the PO. 

\begin{table}[]
    \centering
    % \small
    \resizebox{\columnwidth}{!}{%
    \begin{tabular}{@{}ll@{}}
    \toprule
    \textbf{Feature} & \textbf{Definition} \\ \midrule
    Biomotion & showing movement like that of a living thing \\
    Body & having human or human-like body parts \\
    Human & having human or human-like intentions, plans, or goals \\
    Face & having a human or human-like face \\
    Speech & someone or something that talks \\
    Landmark & having a fixed location, as on a map \\
    Scene & bringing to mind a particular setting or physical location \\ \bottomrule
    \end{tabular}%
    }
    \caption{Feature definitions from \citet{binder_toward_2016}.}
    \label{tab:binder_feats}
\end{table}

\begin{table}
\centering
\resizebox{0.4\columnwidth}{!}{%
\begin{tabular}{@{}lrr@{}}
\toprule
\textbf{Feature} & \textbf{DO} & \textbf{PO} \\ \midrule
Biomotion & \textbf{1.19} & 0.43 \\
Body & \textbf{1.00} & 0.26 \\
Human & \textbf{0.89} & 0.48 \\
Face & \textbf{0.71} & 0.19 \\
Speech & \textbf{0.68} & 0.13 \\
Landmark & 1.83 & \textbf{3.43} \\
Scene & 2.59 & \textbf{4.43} \\ \bottomrule
\end{tabular}%
}
\caption{Relevant Binder features predicted for ``London'' in (1) using CWEs from BERT layer 8. The PO construction lends itself more towards ``location'' features, and the DO more towards animate features.}
\vspace{-1em}
\label{table:london}
\end{table}

To test this phenomenon more robustly, we use a method similar to the experiment for studying grammatical roles in \citet{chronis-etal-2023-method}, which requires a balanced dataset of \textit{DO} and \textit{PO} sentences. While the dataset provided by \citet{hawkins-etal-2020-investigating} is balanced in terms of the two constructions, it is not well-suited to our needs because the variation in recipient animacy is not focused on the place-like versus animate distinction observed in (1). 
Instead, we generate 450 alternating pairs in which the recipient is interpreted by a human evaluator to be a person in the DO and a place in the PO. 
\begin{figure}[!t]
    \centering
    \includegraphics[width=\linewidth]{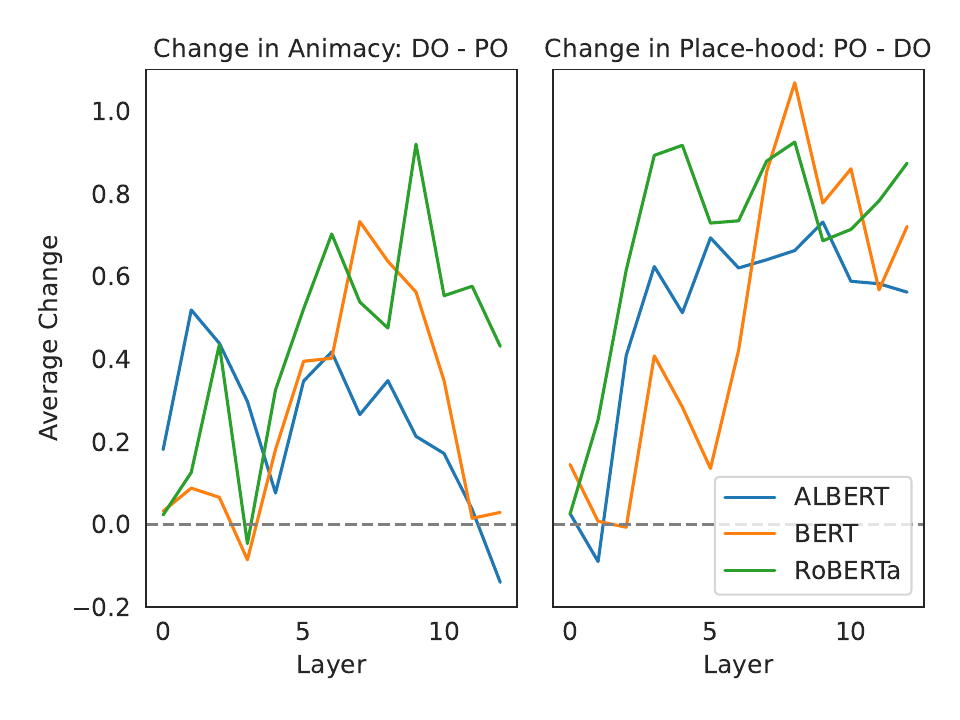}
    \caption{For each layer of an LM, we extract the CWE and project it into Binder space. \textbf{Left:} we measure the average change across the test sentences in Person features from PO to DO. The positive values indicate that recipients in the DO are found to be more animate. \textbf{Right:} we measure the average change across test sentences in Place features from DO to PO. Here, the positive values indicate that the recipients were found to be more place-like in the PO.}
    \label{fig:2panel}
    \vspace{-1em}
\end{figure}
We do this by querying ChatGPT to come up with proper nouns that can be interpreted as places or people, ending up with 15 different such names, all of which were manually checked. 
This included names of states, as in ``Dakota'', and names of countries, as in ``Jordan'', in addition to names of cities, as in ``London''.
We then paired them with 6 different alternating verbs (lemma: send, mail, order, bring, fax, deliver) along with a host of corresponding indirect objects which could also be plausibly received by a place or person. Finally, we choose from five different agents (names), leading to our 450 pairs of sentences, each of the form \texttt{[agent] [verb]$_{past}$ [recipient] [theme]} for DO and \texttt{[agent] [verb]$_{past}$ [theme] to [recipient]} for PO. 
We project the embeddings of the recipients in context from BERT, RoBERTa, and ALBERT to the Binder feature space and average across construction (DO or PO) and feature set (Person or Place). \Cref{fig:2panel} shows the average change in feature values for person-hood features vs. place-hood features across the alternants of the dative construction. That is, a value of 0.75 in the animacy panel (left) suggests that the average difference in the activation value of the recipient's animacy features in the DO and PO constructions was 0.75 
units on a scale of 0 to 6 (as provided by \citet{binder_toward_2016}) with positive values indicating ``more animate in DO than in PO.''
Similar interpretation (though in the reversed direction) can be made for the right panel, which focuses on place-hood change when switching from DO to PO.

\paragraph{Results} As expected, almost all of the models predict an increase in animacy in the DO compared to the PO and an increase in place-hood in the PO compared to the DO (Figure \ref{fig:2panel}) across most layers. There are some exceptions where the change in person-hood/animacy features is in the opposite direction, though these are in the tiny minority (i.e., a total of 3 times out of a total 36 possible model and layer combination). Corroborating with \citet{chronis-etal-2023-method}, we observe particularly high activation-change of the relevant features in layers 6--9 as opposed to the final layer, suggesting possible concentration of semantic sensitivity in those layers. Overall, this suggests that the contextually sensitive distributional semantic embeddings of LMs capture subtle changes in semantic interpretation of different related-constructions.

\section{Conclusion}
Our hope is that both the complete \texttt{semantic-features} library for projecting CWEs into semantic spaces and the online demo will facilitate running linguistically informative experiments using contextual word embeddings.

\section*{Acknowledgments}
Thanks to the three anonymous reviewers for their feedback. We also thank Katrin Erk and Gabriella Chronis for their comments on an earlier draft and the research project in general. (KM)$^2$ acknowledge funding from NSF Grant 2104995 awarded to Kyle Mahowald.

{
\bibliography{anthology,anthology_p2,custom,jwalanthi,kanishka}
}
\appendix

\section{Hyperparameters}
The following search ranges are used by \texttt{optuna} for hidden size, batch size, and learning rate when enabled. 

\begin{table}[!h]
\centering
\resizebox{\columnwidth}{!}{%
\begin{tabular}{@{}lrr@{}}
\toprule
    \textbf{Hyperparameter} & \textbf{Lower Limit} & \textbf{Upper Limit}\\ \midrule
    Hidden Size & \textit{min} & min(2*\textit{min}, \textit{max}) \\
    Batch size & 16 & 128 \\
    Learning Rate & $10^{-6}$ & 1 \\ \bottomrule
\end{tabular}
}
\caption{Search ranges for optimization, where \textit{min} denotes the minimum between the length of the embedding and length of the feature vector and \textit{max} denotes the maximum between the two values}
\label{table:optuna}
\end{table}

\section{Demo Models}
All 117 models available through the Gradio app have 2 layers with 50\% dropout, and early stopping after 6 epochs of non-decreasing validation loss. The maximum epoch limit was set to 100, though in reality, the best performing models finished training after 40-60 epochs. Hyperparameter tuning was used for hidden size, batch size, and learning rate, and pruning was not enabled. For the Buchanan models, the raw feature labels were not used, and the normalized feature values were used. In total, training all 117 models took 25 GPU hours, including those which were discarded in the process of optimization. Models were trained using an NVIDIA A40 GPU.

\end{document}